\documentclass{article}

\usepackage{microtype}
\usepackage{graphicx}
\usepackage{subfigure}
\usepackage{booktabs}

\usepackage{hyperref}

\usepackage[accepted]{icml2024}

\usepackage{amsmath}
\usepackage{amssymb}
\usepackage{mathtools}
\usepackage{amsthm}

\usepackage{amsmath,amsfonts,bm,amssymb,mathtools}

\def\eqref#1{equation~\ref{#1}}

\def\1{\bm{1}}

\def\vu{{\bm{u}}}
\def\vv{{\bm{v}}}

\DeclareMathAlphabet{\mathsfit}{\encodingdefault}{\sfdefault}{m}{sl}
\SetMathAlphabet{\mathsfit}{bold}{\encodingdefault}{\sfdefault}{bx}{n}

\def\gN{{\mathcal{N}}}

\newcommand{\E}{\mathbb{E}}

\newcommand{\KL}{D_{\mathrm{KL}}}

\usepackage[version=3]{mhchem}
\usepackage{balance}

\usepackage[capitalize,noabbrev,nameinlink]{cleveref}

\theoremstyle{plain}

\theoremstyle{definition}

\theoremstyle{remark}

\usepackage{algorithm,algpseudocode}
\usepackage{xpatch}
\makeatletter
\xpatchcmd{\algorithmic}{\ALG@tlm\z@}{\ALG@tlm\z@\leftmargin 0pt}{}{}
\makeatother

\usepackage[textsize=tiny]{todonotes}

\icmltitlerunning{Balancing Molecular Information and Empirical Data 
in the Prediction of Physico-Chemical Properties}

\begin{document}

\twocolumn[
\icmltitle{Balancing Molecular Information and Empirical Data\\
in the Prediction of Physico-Chemical Properties}



\icmlsetsymbol{equal}{*}

\begin{icmlauthorlist}
\icmlauthor{Johannes Zenn}{AIC,UT,IMPRS}
\icmlauthor{Dominik Gond}{RPTU}
\icmlauthor{Fabian Jirasek}{RPTU}
\icmlauthor{Robert Bamler}{AIC,UT}
\end{icmlauthorlist}

\icmlaffiliation{AIC}{AI Center Tübingen}
\icmlaffiliation{UT}{University of Tübingen}
\icmlaffiliation{IMPRS}{IMPRS-IS}
\icmlaffiliation{RPTU}{Laboratory of Engineering Thermodynamics (LTD), RPTU Kaiserslautern}

\icmlcorrespondingauthor{Johannes Zenn}{johannes.zenn@uni-tuebingen.de}

\icmlkeywords{activity coefficients, physico-chemical prediction, machine learning, graph neural networks}

\begin{onecolumnabstract}
  Predicting the physico-chemical properties of pure substances and mixtures is a central task in thermodynamics. Established prediction methods range from fully physics-based ab-initio calculations, which are only feasible for very simple systems, over descriptor-based methods that use some information on the molecules to be modeled together with fitted model parameters (e.g., quantitative-structure-property relationship methods or classical group contribution methods), to representation-learning methods, which may, in extreme cases, completely ignore molecular descriptors and extrapolate only from existing data on the property to be modeled (e.g., matrix completion methods). In this work, we propose a general method for combining molecular descriptors with representation learning using the so-called expectation maximization algorithm from the probabilistic machine-learning literature, which uses uncertainty estimates to trade off between the two approaches.
  The proposed hybrid model exploits chemical structure information using graph neural networks, but it automatically detects cases where 
  structure-based predictions are unreliable, in which case it corrects them by representation-learning based predictions that can better specialize to unusual cases.
  The effectiveness of the proposed method is demonstrated using the prediction of activity coefficients in binary mixtures as an example. The results are compelling, as the method significantly improves predictive accuracy over the current state of the art, showcasing its potential to advance the prediction of physico-chemical properties in general.
\end{onecolumnabstract}

\vskip 0.3in
]



\printAffiliationsAndNotice{}  

\section{Introduction}

Information on physico-chemical properties  is crucial for the conceptual design and optimization of processes in many industries, including chemistry, pharmacy, and biotechnology.
Among the most important thermodynamic properties are the activity coefficients of the components in a mixture, which describe the deviation of a mixture from the ideal mixture and enable the accurate prediction of reaction and phase equilibria of mixtures.
Activity coefficients at infinite dilution are more sensitive thermodynamic properties than activity coefficients at finite concentration (and the subsequently calculated reaction and phase equilibria). Knowing the activity coefficients at infinite dilution allows to predict the activity coefficients in binary mixtures of any finite concentration as well as the activity coefficients in multi-component mixtures.
Unfortunately, measuring thermodynamic properties of mixtures, such as activity coefficients, is costly and time-consuming, and the number of relevant mixtures exceeds the ones that can be studied experimentally by orders of magnitude \citep{jirasek23combining}. 
Consequently, prediction methods for thermodynamic properties of mixtures are paramount.
Recently, research on such prediction methods has split into two branches.

On the one hand, descriptor-based methods correlate information on the molecules to be modeled with properties of interest. Among these, group-contribution methods, which use the composition of the components in terms of structural groups as molecular descriptors and whose underlying equations are usually derived from physical theories, are still the gold standard for property prediction in many (industrial) fields \citep{gmehling15group,jirasek23prediction}. 
The most successful group-contribution method for predicting activity coefficients is UNIFAC \citep{fredenslund1975group,weidlich1987amodified,constantinescu2016further}, which is available in different versions and established in most process simulation software. 
Besides the physics-based group-contribution methods, also other descriptor-based methods that rely on various descriptors, such as molecular weight or surface area, or boiling point,
have been proposed for predicting activity coefficients \cite{katritzky2010quantitative,estrada2006predicting,giralt2004estimation,mitchell1998prediction,paduszynski2016silico,ajmani2008characterization,behrooz2017prediction,medina2022graph}.

In statistics parlance, such descriptor-based models are called \emph{parametric} models as they fit model parameters that affect several related components (e.g., those sharing a given structural group). Thus, parametric models can leverage statistical strength across chemically similar components.

On the other hand, recent idea transfer from the machine learning community has led to an alternative approach to predicting activity coefficients and other mixture properties, which is based solely on learned representations without relying on descriptors.
So-called matrix completion methods (MCMs) \cite{jirasek2020machine,jirasek2020hybridizing,hayer22prediction,grossmann22database,damay21predictin} ignore the chemical structure of components and fit individual representation vectors for each mixture component that appears in a set of available experimental data.
While this approach makes MCMs more flexible than descriptor-based methods, it prevents them from exploiting structural similarities across components, and, therefore, from extrapolating to new components.
In statistics parlance, one says that MCMs are ``\emph{nonparametric} in the components'' (note that nonparametric models tend to have many parameters, similar to how a ``stepless'' controller has infinitely many steps).

It was shown empirically \cite{jirasek2020machine,jirasek2020hybridizing,jirasek2022making} that purely nonparametric MCMs make more accurate predictions for activity coefficients than the descriptor-based (parametric) state-of-the-art UNIFAC.
However, since each fitted parameter (i.e., each representation vector) in an MCM only describes a single component, MCMs can only make predictions for mixtures where each component appears in some (other) mixtures in the available experimental data (``in-domain predictions''). 
By contrast, descriptor-based methods can exploit the structural similarity of components to extrapolate to components that appear in no mixture in the available experimental data (``out-of-domain predictions'').

In this work, we propose a new method for predicting activity coefficients in binary mixtures that combines the strengths of both the parametric (descriptor-based) and the nonparametric (representation-based) approach while avoiding their respective weaknesses.
To do this, we phrase both a parametric and a nonparametric model in a probabilistic framework, and we fit them jointly using the so-called variational expectation maximization (variational EM) algorithm.
The algorithm finds an optimal compromise between the parametric and the nonparametric part of a model, taking into account how confident each part is in its fitted or predicted parameters (discussed in \Cref{sec:method}).
Our evaluation shows that weighing off the respective confidences of the parametric and nonparametric models indeed improves the accuracy of both in-domain and out-of-domain predictions.

While this paper focuses on the concrete task of predicting activity coefficients of pure solutes at infinite dilution in pure solvents at room temperature, the proposed method can, e.g., be generalized to arbitrary temperatures and concentrations following the procedure described in \citet{jirasek2022making}, and to other thermodynamic properties of binary mixtures by fitting it to a corresponding dataset.
More generally, we argue that the variational EM algorithm is a valuable tool in thermodynamic modeling since it allows for combining the strengths of descriptor-based (parametric) and representation-based (nonparametric) models, which is a powerful approach beyond the modeling single thermodynamic properties of binary mixtures.

In the remaining sections of this paper, we first formalize the problem setup, present the proposed method, and discuss several variants of its concrete execution.
We then empirically evaluate the accuracy of predicted activity coefficients and compare them to existing methods and simplified variants of our proposed method (ablation studies).

\section{Method}\label{sec:method}

\begin{figure*}[ht]
    \centering
    \includegraphics[width=0.9\textwidth]{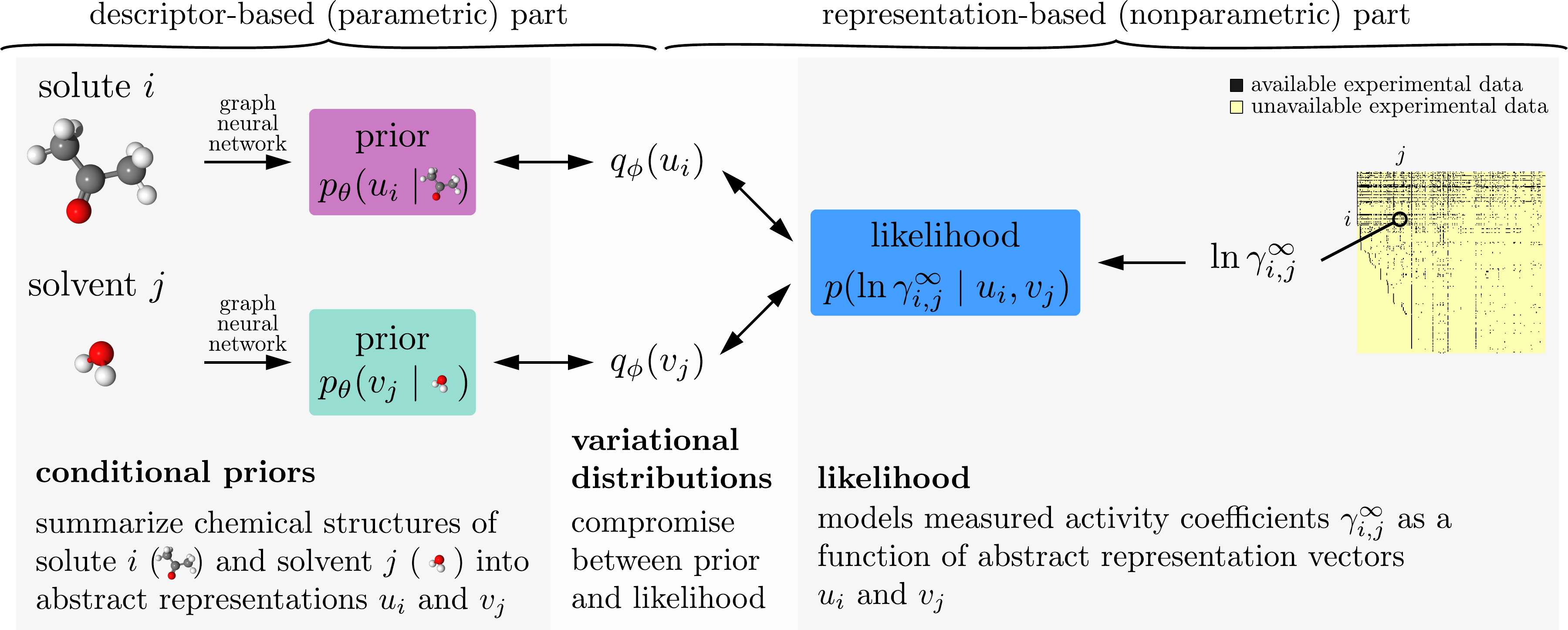}
    \caption{
      Model and data flow for training the proposed model. 
      Left: graph neural networks take chemical structure information and output the parameters of conditional prior probability distributions (\Cref{eq:priors}) over abstract representation vectors.
      Right: the likelihood (\Cref{eq:likelihood}) models how well given representation vectors explain experimentally measured activity coefficients~$\gamma_{i,j}^\infty$.
      We use variational EM (\Cref{sec:vem-intuition,sec:vem-formal}) to fit the neural network weights~$\theta$ (parametric, descriptor-based part), and to find variational distributions for each solute and solvent (nonparametric, representation-based part).
    }
    \label{fig:dataflow-training}
\end{figure*}

\subsection{Problem Setting\label{sec:problem-setting}}

As in \citet{jirasek2020machine}, we start from a data set of $4094$ measured activity coefficients~$\gamma_{i,j}^\infty$ of solutes~$i$ at infinite dilution in solvents~$j$ at $298.15~(\pm1)~\text{K}$. 
We use the same data set as in previous work \cite{jirasek2020machine}, which was taken from the Dortmund Data Bank (DDB) \cite{onken1989thedortmund}, the largest database for phsico-chemical properties covering the most relevant molecular components for technical processes.
We refrain from using synthetic datasets because this only demonstrates how well a method can reproduce an available model and does not results in a practically useful new model. 
The DDB dataset is illustrated in the yellow/black matrix on the right of \Cref{fig:dataflow-training} (which is discussed in more detail in \Cref{sec:prob-model} below).
The matrix has $M=240$ rows and $N=250$ columns corresponding to the $M$~distinct solutes and $N$~distinct solvents that appear in the data set, and each black pixel indicates an available experimental data point~$\gamma_{i,j}^\infty$.
Our goal is to predict activity coefficients for the yellow parts of the matrix (``in-domain predictions''), and to also extend the rows and columns of the matrix, i.e., predict activity coefficients that involve yet unstudied solutes or solvents (``out-of-domain predictions'').

A previous deep-learning-based approach \citep{medina2022graph} addressed this prediction problem with a combination of three neural networks.
The first two networks are so-called graph neural networks (GNNs) that take as input the molecular graph structures of the solute and solvent, respectively, i.e., each atom kind, their hybridizations and formal charges, and the type of bond between each pair of atoms.
The GNNs map the molecular graphs to so-called abstract representation vectors $\vu\in\mathbb R^K$ and $\vv\in\mathbb R^K$, respectively, where the dimension~$K$ is a modeling choice.
The third neural network combines $\vu$ and~$\vv$ and outputs a prediction for the activity coefficient for the respective solute at infinite dilution in the respective solvent.
This existing approach can perform out-of-domain predictions because the neural networks can extrapolate to new molecular structures as long as they share some common substructures with the ones in the training data.
But this approach has the downside that it uses an entirely parametric model, i.e., it is limited by the expressiveness of the neural networks and cannot make any exceptions in case some anomalous components behave very differently than structurally similar components.
Our proposed method, described below, accounts for exceptions with anomalous behavior in a nonparametric way.

\subsection{Probabilistic Model\label{sec:prob-model}}

Like in \citet{medina2022graph}, discussed in \Cref{sec:problem-setting} above, our proposed model has a descriptor-based part (left half of \Cref{fig:dataflow-training}) that processes the chemical structures of the solute and solvent independently using two neural networks (one for solutes and one for solvents), and our main results were also obtained by using GNNs here (similar to the model of \citet{medina2022graph}).
Unlike in the previous work, these neural networks parameterize probabilistic models, i.e., their outputs are not representation vectors $\vu$ and~$\vv$ but instead parameters that define so-called conditional prior probability distributions $p_{\theta}(\vu \,|\, r)$ and $p_{\theta}(\vv \,|\, s)$, respectively.
Here, $\theta$~are the neural network weights, and the bar~``$|$'' denotes conditioning on the chemical structure $r$ and~$s$ of the solute and solvent, respectively.
Specifically, the conditional priors in our empirical analysis are normal distributions,
\begin{align}\label{eq:priors}
  \begin{split}
      \!\!p_{\theta}(\vu \,|\, r) 
      &= 
      \mathcal 
      N\big(
      \vu; 
      \prescript{u}{}{\mu}_{\theta}(r), 
      \text{diag}(\prescript{u}{}{\sigma}_{\theta}^2(r)) 
      \big);
      \\
      \!\!p_{\theta}(\vv \,|\, s) 
      &= 
      \mathcal 
      N\big(
      \vv; 
      \prescript{v}{}{\mu}_{\theta}(s),
      \text{diag}(\prescript{v}{}{\sigma}_{\theta}^2(s)) 
      \big)
  \end{split}
\end{align}
where the means $\prescript{u}{}{\mu}_{\theta}(r), \prescript{v}{}{\mu}_{\theta}(s)\in\mathbb R^K$ and variances $\prescript{u}{}{\sigma}_{\theta}^2(r), \prescript{v}{}{\sigma}_{\theta}^2(s) \in\mathbb R_{>0}^K$ are extracted from the outputs of the two neural networks.
Here, $\text{diag}(\prescript{u/v}{}{\sigma}_{\theta}^2(\,\cdot\,))$ is a covariance matrix with the components of~$\prescript{u/v}{}{\sigma}_{\theta}^2(\,\cdot\,)$ on its diagonal, and zeros on all off-diagonal entries.
The inference algorithm, described in \Cref{sec:vem-intuition} and \Cref{sec:vem-formal} below, ensures that $\prescript{u/v}{}{\sigma}_{\theta}^2(\,\cdot\,)$ estimates an uncertainty region around the corresponding mean prediction $\prescript{u/v}{}{\mu}_{\theta}(\,\cdot\,)$ of the parametric part of the model.
These uncertainty estimates affect how strongly the parametric part of the model constrains (``regularizes'') the nonparametric part of the model during training which we describe next.

The representation-based (nonparametric) part of our model is a probabilistic MCM \cite{jirasek2020machine}.
It represents each solute~$i$ and each solvent~$j$ that appears in the experimental data with an individual representation vector $\vu_i,\vv_j \in \mathbb R^K$, respectively, which it uses to predict the activity coefficients $\gamma_{i,j}^\infty$.
Since activity coefficients range over several orders of magnitude, we model their logarithm, $\ln\gamma_{i,j}^\infty$.
We use a simple Gaussian likelihood,
\begin{align}\label{eq:likelihood}
  p(\ln\gamma_{i,j}^\infty \,|\, \vu_i, \vv_j)
  = \gN(\ln\gamma_{i,j}^\infty; \vu_i \cdot \vv_j, \lambda^2)
\end{align}
where ``$\cdot$'' denotes the dot product and $\lambda = 0.15$ as proposed in previous work \cite{jirasek2020machine}.
While more expressive likelihoods are compatible with our setup, we found the simple choice of \Cref{eq:likelihood} to be sufficient.

\subsection{Fitting the Model: Intuition\label{sec:vem-intuition}}

While the inference algorithm that we use is easy to implement (see \Cref{algo:variational-em} discussed in Section~\ref{sec:vem-formal} below), understanding why it works requires more explanation.
We therefore first motivate the algorithm in this section before formalizing it mathematically in Section~\ref{sec:vem-formal}.

We propose to train the nonparametric and parametric parts of the model jointly using the so-called variational expectation maximization (variational EM) \cite{dempster1977maximum, beal2003variational} algorithm.
Variational EM allows us to fit a model that can generalize across components with similar chemical structures while still being able to learn exceptions for individual components where the experimental data shows evidence for anomalous behavior.

The arrows in \Cref{fig:dataflow-training} show the direction of data flow in the algorithm.
It concurrently fits both the neural network weights~$\theta$ of the conditional priors and a so-called variational distribution $q_\phi(\vu_i)$ and $q_\phi(\vv_j)$ for each solute~$i$ and each solvent~$j$ that appears in the experimental data.
The weights~$\theta$ of the conditional priors are fitted to model the data as well as one can with a parametric model.
By contrast, the variational distributions are fitted in a nonparametric way.
They are fitted to find a compromise between the conditional priors (which can share statistical strength across chemically similar components but cannot make exceptions for anomalous cases) and the experimental data (which may contain evidence for anomalous behaviors, but which is often scarce and generally affected by measurement errors).

\paragraph*{Conceptual Remark on Empirical Bayes Methods.}
Readers who are experienced with Bayesian inference may find it strange that we fit the prior distribution to the data.
In normal Bayesian inference, one seeks the posterior distribution of some experimental data under a given probabilistic model, and one assumes that the prior of the probabilistic model is given (e.g., informed by expert knowledge).
Variational EM falls into the class of so-called empirical Bayes methods, which differ from normal Bayesian inference in that they estimate the prior distribution from the data as well.
This would be an underspecified problem if the prior was unconstrained, in which case the prior would overfit to the data, and the resulting posterior would be equal to the prior and thus also overfit, i.e., perfectly explain the available data but fail to generalize beyond it.
To avoid this collapse of empirical Bayes, one has to constrain the prior to a smaller class of distributions than the posterior.

In our setup, the necessary constraint on the prior comes from the finite expressiveness of the neural networks: unless the neural network for, e.g., solute representation vectors~$\vu_i$ is exorbitantly large, it cannot output completely independent prior parameters $\big(\prescript{u}{}{\mu}_{\theta}(r_i), \prescript{u}{}{\sigma}_{\theta}^2(r_i)\big)$ for all solutes~$i$ in the dataset.
Thus, fitting the neural network weights~$\theta$ cannot perfectly overfit the prior to the data.
By contrast, the variational distributions $q_\phi(\vu_i)$ and $q_\phi(\vv_j)$ are fitted nonparametrically, i.e., with individual parameters for each solute and solvent.
The reason why these do not perfectly overfit the data is because they are not fitted solely to the data but instead obtained by (approximate) Bayesian inference with the (non-overfitting) prior (explained in \Cref{sec:vem-formal}).

\begin{figure}[!t]
  \centering
  \includegraphics[width=\columnwidth]{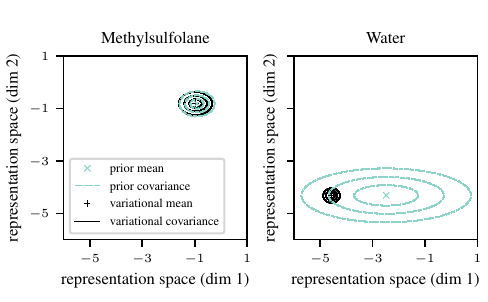}
  \caption{
      Influence of prior uncertainty estimates (turquoise) on the final fitted parameters (black) for methylsulfolane (least frequent solvent, left) and water (most frequent solvent, right).
      Concentric ellipses show 25\%, 75\%, and 95\% quantiles, respectively.
      For low prior uncertainty (small turquoise ellipses, left), the final fit is forced to closely match the prior, while a large prior uncertainty (right) admits more freedom to the final fit.
      Discussion in \Cref{sec:vem-intuition} and model architectures in \Cref{sec:evaluation-setup}.
  }
  \label{fig:compromise}
\end{figure}

\paragraph*{The role of prior uncertainties.}
Our ablation studies in the results section show that it is indeed crucial that the compromise between the parametric and the nonparametric fit takes the uncertainty estimates $\prescript{u}{}{\sigma}_{\theta}^2(r)$ and $\prescript{v}{}{\sigma}_{\theta}^2(s)$ of the conditional priors into account.
\Cref{fig:compromise} shows two examples of how prior uncertainties affect the training in variational EM.
The two panels show 2-dimensional cuts of the representation spaces for the solvents methylsulfolane (left) and water (right).
We picked the two dimensions in representation space in which prior and variational distribution differ most (measured by Kullback-Leibler divergence \citep{kullback1951information, murphy2022probabilistic}).
The dashed turquoise and solid black ellipses show 25\%, 75\%, and 95\% quantiles of the conditional priors (\Cref{eq:priors}) and variational distributions, respectively.

The positions of the ellipses in representation space are not directly interpretable, but their sizes indicate uncertainty estimates.
For example, the prior predictions for methylsulfolane (left panel in \Cref{fig:compromise}) have low uncertainty (the turquoise ellipses are small).
This is expected to happen for solvents (and equally for solutes) where the dataset contains structurally similar solvents that empirically behave similarly in mixtures, thus allowing the neural network to effectively and confidently interpolate between them.
The low prior uncertainty causes variational EM to trust the prior predictions, and to fit a variational distribution $q_\phi(\vv_j)$ (solid black ellipses) that closely follows the prior.

By contrast, the prior predictions for water (right panel in \Cref{fig:compromise}) have high uncertainty (the turquoise ellipses are large).
This is expected to happen if the dataset contains solvents that are structurally similar to water but behave very differently in mixtures.
Such anomalous cases prevent the neural network from interpolating effectively.
However, as we discuss in \Cref{sec:vem-formal} below, the neural network is at least fitted to detect such cases, and to reflect them by outputting a large uncertainty estimate~$\prescript{v}{}{\sigma}_{\theta}^2(s)$.
As can be seen in the right panel of \Cref{fig:compromise}, the high prior uncertainty allows variational EM to fit the variational distribution (black ellipses) more freely, thus, in a sense, overriding the descriptor-based prior (mean) prediction $\prescript{v}{}{\mu}_{\theta}(s)$ (the turquoise cross) in this case.
Note that the uncertainty of the (approximate) \emph{posterior} (the size of the black ellipses) for water is small despite the large prior uncertainty.
This is expected since the dataset contains a lot of experimental data where the solvent is water.

\subsection{Inference Algorithm\label{sec:vem-formal}}

We now formally discuss the variational EM algorithm \cite{dempster1977maximum, beal2003variational} in the concrete context of the model for activity coefficients in binary mixtures which has been introduced in \Cref{sec:prob-model}.
Combining the conditional priors (\Cref{eq:priors}) and the likelihood (\Cref{eq:likelihood}), our probabilistic model defines a joint probability density over all representation vectors $\vu_i$ and $\vv_j$, and all logarithmic activity coefficients $\ln\gamma_{i,j}^\infty$ in all binary mixtures ${i{-}j}$ in the dataset,
\begin{align}\label{eq:joint}
  \begin{split}
    &p_{\theta}(\bm u,\bm v,\ln \bm\gamma^\infty \,|\, \bm r,\bm s)
      =
      \Big(\prod_{i=1}^M p_\theta(u_i\,|\,r_i) \Big) \times \\
      &\quad\times \Big(\prod_{j=1}^N p_\theta(v_j\,|\,s_j) \Big) \times 
      \Big(\!\!\prod_{(i,j)\in\mathcal D} \!\!\! p(\ln \gamma_{i,j}^\infty \,|\, u_i, v_j) \Big).
  \end{split}
\end{align}
Here, our notation of boldface symbols $\bm u$, $\bm v$, $\bm r$, $\bm s$, and $\ln\bm\gamma^\infty$ on the left-hand side denotes the collection of all representation vectors~$u_i$ and $v_j$ and all chemical structures~$r_i$ and $s_j$ for all solutes~$i$ and all solvents~$j$, respectively, that appear at least once in the experimental data~$\mathcal D$, and all logarithmic activity coefficients $\ln\gamma_{i,j}^\infty$ of all binary mixtures ${i{-}j}$ for which experimental data is available.
Similarly, the first two products on the right-hand side of \Cref{eq:joint} run over all $M$ solutes~$i$ and all $N$ solvents~$j$, respectively, and the third product runs over all pairs $(i,j)$ where we have experimental data for the binary mixture ${i{-}j}$ (i.e., the black pixels in the yellow/black matrix on the right of \Cref{fig:dataflow-training}).

A naive approach to training the neural networks would attempt to find the network weights~$\theta$ that maximize the so-called marginal likelihood $p_{\theta}({\ln\bm\gamma^\infty\,|\,\bm r,\bm s})$ which is the probability density of predicting the experimentally measured logarithmic activity coefficients $\ln\bm\gamma^\infty$ for all binary systems that are contained in the available experimental dataset.
Unfortunately, the marginal likelihood is not accessible in our model because obtaining it would require marginalizing \Cref{eq:joint} over $\bm u$ and~$\bm v$,
\begin{align}\label{eq:marginal-likelihood}
  p_{\theta}({\ln\bm\gamma^\infty \;\!|\;\! \bm r,\bm s})
  \!=\!\! \int\!\!\!\!\!\int\!\! p_{\theta}(\bm u,\bm v,\ln \bm\gamma^\infty \;\!|\;\! \bm r,\bm s) \;\!\text{d}\bm u \;\!\text{d}\bm v
\end{align}
which is a high-dimensional integral that is prohibitively computationally expensive to calculate.
Variational EM instead resorts to an approximate method called variational inference \citep{blei2017variational,zhang2018advances}, which provides a lower bound on the log marginal likelihood, called the evidence lower bound (ELBO),
\begin{align}\label{eq:elbo-bound}
  \operatorname{ELBO}(\theta,\phi) \leq \ln p_{\theta}({\ln\bm\gamma^\infty \,|\, \bm r,\bm s}) \quad\forall \theta,\phi.
\end{align} 
Here, $\phi$ are the so-called variational parameters.
We discuss $\phi$ and define the ELBO below.
The ELBO is useful because---unlike the marginal likelihood---it can be estimated efficiently, and maximizing it over both $\theta$ and~$\phi$ serves as a proxy for maximizing the log marginal likelihood on the right-hand side of \Cref{eq:elbo-bound}:
since the bound in \Cref{eq:elbo-bound} holds for all values of~$\phi$, and $\phi$ only appears on the left-hand side, maximizing the ELBO over~$\phi$ makes the bound as tight as possible.
Maximizing the ELBO also over~$\theta$ thus finds neural network weights for which we can at least give the best guarantee for the marginal likelihood.

To derive a valid expression for the ELBO, variational inference replaces the integral on the right-hand side of \Cref{eq:marginal-likelihood} with a form of biased importance sampling \cite{bamler2017perturbative}.
One first chooses a family of typically simple probability distributions $q_\phi(\bm u,\bm v)$ that are parameterized by~$\phi$ and called variational distributions.
For simplicity, we use the so-called Gaussian mean-field approximation, i.e., we choose a family of fully factorized normal distributions $q_\phi(\bm u,\bm v) = \big(\prod_{i=1}^M q_\phi(u_i)\big) \big(\prod_{j=1}^N q_\phi(v_j)\big)$ with
\begin{align}\label{eq:var-dist}
  \begin{split}
    q_\phi(\vu_i) &= \mathcal N\big(\vu_i; \prescript{u}{}{\tilde\mu}_i, \operatorname{diag}(\prescript{u}{}{\tilde\sigma}_i^2)\big); \\
    q_\phi(\vv_j) &= \mathcal N\big(\vv_j; \prescript{v}{}{\tilde\mu}_j, \operatorname{diag}(\prescript{v}{}{\tilde\sigma}_j^2)\big)
  \end{split}
\end{align}
where the variational means $\prescript{u}{}{\tilde\mu}_i, \prescript{v}{}{\tilde\mu}_j \in \mathbb R^K$ and variances $\prescript{u}{}{\tilde\sigma}^2_i, \prescript{v}{}{\tilde\sigma}^2_j \in \mathbb R_{>0}^K$ together make up the variational parameters~$\phi$.
The ELBO is then \cite{blei2017variational}
\begin{align}\label{eq:elbo}
    \!\!\!\!\!\!\!\!\operatorname{ELBO}(\theta,\phi)
    &= \sum_{\!\!\!\!\!(i,j)\in\mathcal D\!\!\!\!\!} \E_{q_\phi(\vu_i)\,q_\phi(\vv_j)}\big[\ln p(\ln \gamma_{i,j}^\infty \,|\, \vu_i, \vv_j)\big] \nonumber\\
    &\;\;\;\;\, - \sum_i \KL\big(q_\phi(\vu_i) \,\big\|\, p_\theta(\vu_i \,|\, r_i)\big) \\
    &\;\;\;\;\, - \sum_j \KL\big(q_\phi(\vv_j) \,\big\|\, p_\theta(\vv_j \,|\, s_j)\big) \nonumber.
\end{align}
Here, the first term in the sum is the expectation value~$\E[\,\cdot\,]$ of the log likelihood under the variational distribution, which can be estimated by averaging the logarithm of \Cref{eq:likelihood} over samples $u_i\sim q_\phi(u_i), v_j\sim q_\phi(v_j)$.
The following two terms in the sum are Kullback-Leibler (KL) divergences \citep{kullback1951information, murphy2022probabilistic}, which quantify how much the variational distributions differ from the conditional priors.
For normal distributions, the KL divergence can be calculated analytically \cite{murphy2022probabilistic}
\begin{align}\label{eq:kl-gaussian}
    \KL&\big(q_\phi(\vu_i) \,\big\|\, p_\theta(\vu_i \,|\, r_i)\big)\nonumber\\
    &=\frac12 \sum_{\alpha=1}^K \bigg[
        \frac{\big(\!\prescript{u}{}{\tilde\mu}_{i,\alpha} - \prescript{u}{}{\mu}_\theta(r_i)_\alpha\big)^2}{\prescript{u}{}{\sigma^2_\theta}(r_i)_\alpha}
        + \frac{\prescript{u}{}{\tilde\sigma^2_{i,\alpha}}}{\prescript{u}{}{\sigma^2_\theta}(r_i)_\alpha} \\
    &\qquad\qquad + \ln\big(\prescript{u}{}{\sigma^2_\theta}(r_i)_\alpha\big)
        - \ln\big(\prescript{u}{}{\tilde\sigma}^2_{i,\alpha}\big)
        - 1
    \bigg]\nonumber
\end{align}
(analogously for $\KL\big(q_\phi(\vv_j) \,\big\|\, p_\theta(\vv_j \,|\, s_j)\big)$), where $\alpha$ indexes the coordinate in $K$-dimensional representation space.

Maximizing the ELBO in \Cref{eq:elbo} over both $\theta$ and~$\phi$ trades off between three objectives:
\begin{enumerate}
  \item[(i)]
    maximizing the first term on the r.h.s.\ of \Cref{eq:elbo} over~$\phi$ tries to fit variational distributions $q_\phi(u_i)$ and $q_\phi(v_j)$ in a way that samples from these distributions explain the experimental data in~$\mathcal D$;
  \item[(ii)]
    maximizing the last two terms in \Cref{eq:elbo} over~$\phi$ (which amounts to minimizing the KL-divergences over~$\phi$) regularizes the fits, i.e., it keeps the variational distributions $q_\phi(u_i)$ and $q_\phi(v_j)$ close to the conditional priors.
    Here, the first term on the r.h.s.\ of \Cref{eq:kl-gaussian} penalizes deviations between prior mean and variational mean stronger for smaller prior variance~$\prescript{u}{}{\sigma^2_\theta}(r_i)_\alpha$.
    Thus, the (parametric) prior model has a stronger effect on the (nonparametrically fitted) variational distributions when it is confident in its prediction, as claimed in the discussion of \Cref{fig:compromise};
  \item[(iii)]
    minimizing the KL-divergences in \Cref{eq:elbo} also over~$\theta$ fits the neural networks that define the conditional priors to the variational distributions, and thus indirectly to the data.
    This includes fitting the prior variances~$\prescript{u/v}{}{\sigma^2_\theta}(\cdot)$ to model the aleatoric uncertainty observed in the data plus any changes between the variational distributions of structurally similar components that cannot be resolved by the prior due to the finite expressiveness of the neural networks.
\end{enumerate}

We maximize the ELBO over $\theta$ and~$\phi$ with stochastic gradient descent, using reparameterization gradients \cite{kingma2013auto} for the first term on the right-hand side of \Cref{eq:elbo}, and automatic differentiation provided by common software frameworks for machine learning \cite{pytorch}.
\Cref{algo:variational-em} summarizes the algorithm.
Our implementation is available online (see section ``Data and Software Availability'').
Training our largest model variant (GNN MCM, see below) took about four hours on a single GPU (Nvidia GeForce RTX 2080 Ti).

\begin{algorithm}[t]
    \caption{Variational Expectation Maximization for GNN MCM}
    \label{algo:variational-em}
    \begin{algorithmic}
        \State {\bfseries Input:} dataset~$\mathcal D$ of activity coefficients~$\gamma_{i,j}^\infty$ in binary mixtures, involving $M$ distinct solutes $i$ and $N$ distinct solvents $j$; model~$p_\theta$ as defined in \Cref{eq:priors,eq:likelihood}; variational family~$q_\phi$ as defined in \Cref{eq:var-dist}; dimension~$K$ of the abstract representation space; learning rate~$\alpha$; size~$m$ of minibatches.
        \vspace{.2em}
        \State {\bfseries Output:} optimized parameters $\theta$ and~$\phi$.
        \vspace{.8em}
        \State Initialize $\theta$ and $\phi\equiv\big((\prescript{u}{}{\tilde\mu}_i, \prescript{u}{}{\tilde\sigma}_i)_{i=1}^M, (\prescript{v}{}{\tilde\mu}_j, \prescript{v}{}{\tilde\sigma}_j)_{j=1}^N\big)$ randomly.
        \vspace{.1em}
        \Repeat
        \State Draw a minibatch~$\mathcal{B}$ of $m$~index pairs~$(i,j)$ for which
          \State\quad experimental data $\gamma_{i,j}^\infty$ exists in~$\mathcal{D}$.
        \vspace{.2em}
        \State Set $\mathcal I \gets \{i: (i,j) \in \mathcal B\}$ and $\mathcal J \gets \{j: (i,j) \in \mathcal B\}$.
        \vspace{.2em}
        \State Draw standard normal noise $\prescript{u}{}{\epsilon}_i \sim \mathcal N(0,I_{K\times K})\, \forall i\in\mathcal I$ 
          \State\quad and set $u_i \gets \prescript{u}{}{\tilde\mu}_i + \prescript{u}{}{\tilde\sigma}_i \odot \prescript{u}{}{\epsilon}_i\;\; \forall i\in\mathcal I$.
          \State\quad\emph{(``$\odot\!$'' denotes elementwise multiplication.)}
        \State Draw standard normal noise $\prescript{v}{}{\epsilon}_j \sim \mathcal N(0,I_{K\times K})\, \forall j\!\in\!\mathcal J$ 
          \State\quad and set $v_j \gets \prescript{v}{}{\tilde\mu}_j + \prescript{v}{}{\tilde\sigma}_j \odot \prescript{v}{}{\epsilon}_j\;\; \forall j\in\mathcal J$.
        \vspace{.2em}
        \State Set $\prescript{\gamma}{}{\mathcal L} \gets \frac{|\mathcal D|}{m} \sum_{(i,j) \in \mathcal{B}}\ln p(\ln \gamma_{i,j}^\infty \,|\, u_i, v_j)$.\hspace{\fill}
        \State\quad\emph{$\vartriangleright$ see \Cref{eq:likelihood}}\vspace{.2em}
        \State Set $\prescript{u}{}{\mathcal L} \gets \frac{M}{|\mathcal I|} \sum_{i\in \mathcal I}
        \KL\big(q_\phi(\vu_i) \,\big\|\, p_\theta(\vu_i \,|\, r_i)\big)$.\hspace{\fill}
        \State\quad\emph{$\vartriangleright$ see \Cref{eq:kl-gaussian}}\vspace{.2em}
        \State Set $\prescript{v}{}{\mathcal L} \gets \frac{N}{|\mathcal J|} \sum_{j \in \mathcal J} \KL\big(q_\phi(\vv_j) \,\big\|\, p_\theta(\vv_j \,|\, s_j)\big)$.
        \vspace{.3em}
        \State Set $\operatorname{ELBO}_\mathcal{B}(\theta,\phi) \gets \prescript{\gamma}{}{\mathcal L} + \prescript{u}{}{\mathcal L} +  \prescript{v}{}{\mathcal L}$.
        \vspace{.2em}
        \State Compute gradients $\nabla_{\!\theta} \operatorname{ELBO}_\mathcal{B}(\theta,\phi), \nabla_{\!\phi} \operatorname{ELBO}_\mathcal{B}(\theta,\phi)$ 
        \State\quad using automatic differentiation.
        \vspace{.2em}
        \State Update $\theta \gets \theta + \alpha\nabla_{\!\theta} \operatorname{ELBO}_\mathcal{B}(\theta,\phi)$.
        \vspace{.2em}
        \State Update $\phi \gets \phi + \alpha\nabla_{\!\phi} \operatorname{ELBO}_\mathcal{B}(\theta,\phi)$.
        \vspace{.1em}
        \Until{convergence.}
    \end{algorithmic}
\end{algorithm}

\begin{figure*}[ht]
  \centering
  \includegraphics[width=0.9\textwidth]{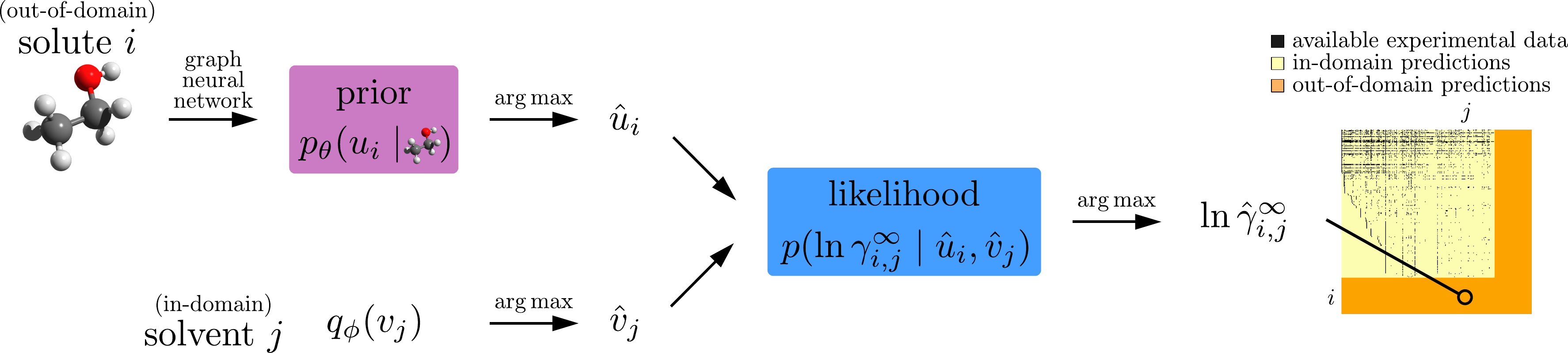}
  \caption{
    Data flow for a prediction where the solvent appears in the training set (in-domain) but the solute does not (out-of-domain).
    We thus predict the solute representation vector~$\hat u_i$ from the prior, and the solvent representation vector~$\hat v_j$ from the variational distribution~$q_\phi$, see \Cref{eq:prediction}.
  }
  \label{fig:dataflow-prediction}
\end{figure*}

\subsection{Predictions}

Once our model is trained with variational EM, we use it for predicting activity coefficients for binary mixtures whose components can each be either in-domain (i.e., appearing in other mixtures in the available experimental data) or out-of-domain (i.e., previously unstudied components).
\Cref{fig:dataflow-prediction} shows an example where the solute~$i$ is out-of-domain whereas the solvent~$j$ is in-domain.
For the out-of-domain solute~$i$, we apply the trained neural network to its chemical structure~$r_i$, which outputs the means and variances of the conditional prior $p_\theta({u_i \,|\, r_i})$ (\Cref{eq:priors}).
For the in-domain solvent~$j$, we directly use the variational distribution $q_\phi(v_j)$ (\Cref{eq:var-dist}), which was fitted to the data under consideration of its conditional prior.
We then obtain a prediction $\hat\gamma_{i,j}^\infty = \exp(\ln\hat\gamma_{i,j}^\infty)$ by calculating the modes,
\begin{align}
    \hat u_i &\coloneqq \arg\max_{u_i} p_\theta({u_i \,|\, r_i}) = \prescript{u}{}{\mu}_{\theta}(r_i); \nonumber\\
    \hat v_j &\coloneqq \arg\max_{v_j} q_\phi(v_j) = \prescript{v}{}{\tilde\mu_j}; \label{eq:prediction}\\
    \ln\hat\gamma_{i,j}^\infty &\coloneqq \arg\max_{\ln\gamma_{i,j}^\infty} p(\ln\gamma_{i,j}^\infty \,|\, \hat u_i,\hat v_j) = \hat u_i \cdot \hat v_j; \nonumber
\end{align}
where $p(\ln\gamma_{i,j}^\infty \,|\, \hat u_i,\hat v_j)$ is the likelihood (\Cref{eq:likelihood}).
For different combinations of in-domain and out-of-domain mixture components, we adapt \Cref{eq:prediction} accordingly.

\subsection{Model Details And Variants}

In our experiments, we investigate two model variants that differ in how they represent chemical information of solutes and solvents, and how they parameterize the means and variances of the conditional prior distributions (\Cref{eq:priors}) as functions of these chemical structures.
A simple model variant, which we call ``MoFo MCM'', represents chemical structures by the molecular formula (MoFo) (e.g., water is represented as \ce{H2O}).
A more expressive model variant, which we call ``GNN MCM'', represents chemical structures by their topological molecular graphs (e.g., water is represented as the graph \hbox{\ce{H}$-$\ce{O}$-$\ce{H}}), and the model employs graph neural networks (GNNs) \cite{gori2005new,scarselli2008computational,bronstein2021geometric}.

In detail, the MoFo MCM uses two neural networks (one for solutes and one for solvents) that receive a fixed-size integer-valued vector as input.
Each entry of the input vector corresponds to a given atom or bond type, and the values at these entries count the number of occurrences of the given atom or bond type in the molecule.
Specifically, we use 16-dimensional input vectors for the 12 atoms \ce{O}, \ce{Si}, \ce{I}, \ce{F}, \ce{Br}, \ce{P}, \ce{H}, \ce{S}, \ce{Sn}, \ce{N}, \ce{C}, and \ce{Cl} present in the dataset and the 4 bond types single, double, triple, and aromatic.
The network outputs a $2K$-dimensional vector that is the concatenation of the prior means $\prescript{u/v}{}{\mu}_{\theta}(\cdot)$ and variances $\prescript{u/v}{}{\sigma}_{\theta}^2(\cdot)$.

The GNN MCM uses two graph neural networks, whose inputs are the molecular graphs of the solute and solvent, respectively.
More specifically, we encode atoms and bonds from the same vocabulary as in the MoFo MCM with learnable real-valued vectors, which we use as initial node and edge features for the GNN.
In general, message-passing GNNs operate on such graph-structured inputs by performing transformations of the node and edge features over multiple layers via a message-passing scheme \cite{gilmer2017neural}. 
The output of a GNN is computed from all node features (and possibly edge features) at the last layer with a readout function. 
Generally, the message-passing scheme consists of a message step, an aggregation step, and an update step.
In each layer, a message is computed for each directed edge utilizing a message function whose parameters are part of the learnable neural network parameters~$\theta$.
Incoming messages are aggregated by a sum for each node. 
The update function produces new node features depending on the previous node features and the aggregated message, and its parameters are also part of~$\theta$.

Message, aggregation, and update steps are specific to the architecture of the GNN.
In this work, we utilize the Feature-wise Linear Modulation GNN \citep{brockschmidt2020film}.
This model uses the target features of a directed edge as input to a hyper-network that determines element-wise affine transformation parameters.
Messages are computed by scaling and shifting the input features with the element-wise affine transformation parameters, where the input features result from multiplying a learnable matrix with previous features.
The update function of a node sums over the aggregated messages for each edge type, where also transformation parameters are computed for each edge type.

\section{Evaluation Setup\label{sec:evaluation-setup}}

In \Cref{sec:results} below, we compare the two variants of our proposed method (``MoFo MCM'' and ``GNN MCM'') to other existing prediction methods (which are called ``baselines'') and to simplified variants of our models that have parts removed (which are called ``ablations'').

\paragraph*{Models And Baselines.}
We evaluate the two variants ``MoFo MCM'' and ``GNN MCM'' of our proposed method.
As baselines, we compare to 
the group-contribution method modified UNIFAC (Dortmund) \citep{weidlich1987amodified,constantinescu2016further} (which we will refer to simply as ``UNIFAC'' in the following) and to two machine-learning based methods: the fully nonparametric MCM method by \citet{jirasek2020machine} and the fully parametric neural-network-based method by \citet{medina2022graph}.
The latter uses a different GNN architecture \citep{gilmer2017neural} than our conditional priors, and it includes more chemical information in the prediction (such as orbital hybridizations and formal atom charges).

\paragraph*{Ablation Studies.}
We perform two ablation studies where we remove parts of our method to investigate their contribution to the method's performance.
For the first ablation study, we use the same trained MoFo MCM and GNN MCM models as for our main results, but we perform predictions for in-domain components as if they were out of domain, i.e., using the mode of the conditional prior as representation vector (see first line of 
\Cref{eq:prediction}), thus ignoring the variational distributions at prediction time.

For the second ablation study, we simplify the model by removing its nonparametric part, and we train it by maximum likelihood estimation (MLE) rather than variational EM.
Thus, in this ablation, the neural networks only output means $\prescript{u}{}{\mu}_{\theta}(r_i)$ and $\prescript{v}{}{\mu}_{\theta}(s_j)$ and no variances, and we use these means directly as representation vectors $u_i$ and~$v_j$, respectively, in the likelihood (\Cref{eq:likelihood}), which our training objective maximizes over the neural network parameters~$\theta$ (similar to \citet{medina2022graph}).
Since there are no variational distributions, predictions are again done as if all mixture components were out of domain.

\paragraph*{Training.}
We train our models with $10$-fold cross validation \cite{goodfellow2016deep}.
For each split, we use $80\%$ of the dataset for training, $10\%$ for testing, and $10\%$ for a validation set.
The 10 resulting test set splits are pairwise disjoint, and their union equals the full dataset. 
We use the test set splits to evaluate the accuracy of model predictions, where we consider a mixture ``${i{-}j}$'' in the test set to be out-of-domain if at least one of solute~$i$ or solvent~$j$ does not appear in the corresponding training split.
Note that our $10$-fold cross validation is different from the work of \citet{jirasek2020machine}, which uses more computationally expensive leave-one-out cross validation.

We implement our models in PyTorch \citep{pytorch} and use PyTorch Geometric \citep{pytorchgeometric} for the GNN.
All models are trained for ${15,000}$ epochs using the Adam \cite{kingma2015adam} optimizer. 
In the MoFo (MLE) ablation study, we use early stopping \cite{morgan1989generalization,zhang2023dive}, i.e., we compute validation errors every $10$~epochs and use the model with the lowest validation mean squared error (MSE) to compute evaluation errors on the test set.
This is done to be as lenient as possible to the ablation study, and because MLE training is more prone to overfitting than variational EM.
When training with variational EM, we do use the validation set. 

To find well-performing hyperparameters (e.g., the learning rate schedule and the dimension~$K$ of the abstract representation space), we utilize a sparse random grid search. 
We provide more information on this process in the supplementary information.
We choose the best model of the grid search according to its MSE on a predefined dataset split that is the same for all models and different from any other split. 
In order to fairly compare all models, we exclude the test data of the predefined dataset (that is used to determine the hyperparameters) from the evaluation.

\citet{medina2022graph} use a different dataset and train an ensemble of  $30$ models for prediction where each model has been trained on randomized train/validation splits.
For a fair comparison against this baseline, we train a separate GNN MCM for each of these train/validation splits, using again a sparse random grid search for hyperparameter tuning.

\begin{figure}[t]
  \centering
  \includegraphics[width=\columnwidth]{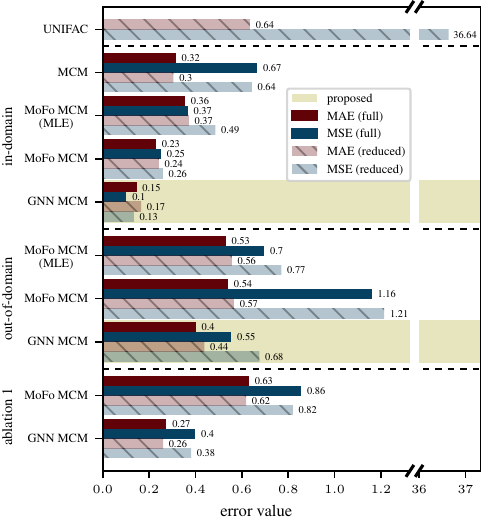}
  \caption{
  In-domain prediction errors (upper part), out-of-domain prediction errors (middle part), and ablations (lower section).
  The ``reduced'' dataset (light hatched bars) contains only mixtures to which UNIFAC is applicable. 
  The proposed GNN MCM (gold highlighting) has the best predictive accuracy for both in-domain and out-of-domain predictions.
  Results labeled ``ablation~1'' (lower section) show errors for in-domain prediction tasks only, but performed as if these were out of domain.
  }
  \label{fig:main-results}
  \vskip-0.5em
\end{figure}

\section{Results and Discussion\label{sec:results}}

\Cref{fig:main-results} and \Cref{tab:comparison-medina} summarize our results by showing the mean absolute error (MAE) and mean squared error (MSE) of the predicted logarithmic activity coefficients of all evaluated models.
Solid bars in \Cref{fig:main-results} show evaluations based on our full dataset.
As the UNIFAC baseline cannot be applied to all mixtures in our dataset, we also trained and evaluated all models on a reduced dataset, which contains only those data points that can be modeled by UNIFAC (light hatched bars).
The neural-network baseline method by \citet{medina2022graph} uses yet a different dataset.
Therefore, we trained and evaluated an additional instance of our proposed GNN MCM method on their dataset, and we compare its predictive accuracy to the results reported by the authors of \citet{medina2022graph} in \Cref{tab:comparison-medina}.
In the following, we discuss all results in detail.

\paragraph*{Comparison to Baselines.}
The proposed GNN MCM (highlighted in gold in \Cref{fig:main-results}) provides more accurate predictions than all considered baselines, both in terms of MAE and MSE, and for both in-domain and out-of-domain predictions.
Compared to UNIFAC \citep{weidlich1987amodified,constantinescu2016further} (first row in \Cref{fig:main-results}), the GNN MCM makes significantly more accurate predictions even if we restrict the test set for GNN MCM to the more difficult out-of-domain predictions (eighth row in \Cref{fig:main-results}).
Predictive accuracy is further improved significantly for in-domain predictions (fifth row in \Cref{fig:main-results}).
Recall that, even for an in-domain prediction, the training data never contains the precise mixture ${i{-}j}$ for which we make a prediction;
it only contains other mixtures that may involve either solute~$i$ or solvent~$j$ or neither, but never both. This is in sharp contrast to UNIFAC, whose training set has not been disclosed. However, one must assume that a significant share of the dataset considered in this work was used to fit UNIFAC, so the UNIFAC results should rather be seen as correlations than predictions, making the performance of the proposed GNN MCM even more impressive.

\begin{table}[t]
  \centering
  \renewcommand{\arraystretch}{1.2}
  \caption{
  Comparison of the proposed GNN MCM with the model from We \citet{medina2022graph} in terms of MAE and MSE. We use the same dataset and splits as \citet{medina2022graph}.
  The table shows mean and standard deviations over $30$ splits. 
  }
  \vskip1em
  \begin{tabular}{l|cc}
       model & MAE & MSE \\
       \hline
       GNN MCM 
       & $\mathbf{0.1542_{\pm 0.0046}}$ 
       & $\mathbf{0.0905_{\pm 0.0071}}$ 
       \\
       \citet{medina2022graph} 
       & $0.1973_{\pm 0.0067}$
       & $0.1196_{\pm 0.0074}$ 
  \end{tabular}
  \label{tab:comparison-medina}
\end{table}

The comparison to the fully nonparametric MCM \citep{jirasek2020machine} (second row in \Cref{fig:main-results}) is only possible for in-domain predictions as this baseline cannot perform out-of-domain predictions.
Here, the GNN MCM (fifth row) approximately halves MAE, and it reduces MSE (which is more sensitive to outliers) even more significantly.

The published evaluation results in \citet{medina2022graph} do not distinguish between in-domain and out-of-domain predictions, effectively averaging over both.
Using the same training data and evaluation setup, our proposed GNN MCM significantly reduces both MAE and MSE (\Cref{tab:comparison-medina}).

The fourth and seventh rows in \Cref{fig:main-results} show prediction errors of the MoFo MCM variant of our model, whose conditional priors only utilize the molecular formula of mixture components but not on their chemical structures.
We find that this model variant performs worse than the GNN MCM model on both in-domain and out-of-domain prediction tasks.
For in-domain predictions, we can compare again to the fully nonparametric MCM (second row in \Cref{fig:main-results}), which does not exploit any chemical information about the mixture components.
We find, as expected, that performance improves with increasing granularity of exploited chemical information: MoFo MCM performs better than the fully nonparametric MCM but worse than GNN MCM.

\paragraph*{Ablation~1: Predicting Without the Nonparametric Model Part.}
Our first ablation discards the nonparametric part of the model after training and performs predictions for in-domain mixtures as if they were out-of-domain (i.e., only using the conditional priors).
The last two rows in \Cref{fig:main-results} show prediction errors of the MoFo MCM model and the GNN MCM model for this ablation.
Here, we evaluate on the same dataset splits as in the in-domain predictions since the splits for out-of-domain predictions contain tasks where this ablation study would not change anything.
Comparing the last two rows of \Cref{fig:main-results} to rows four and five, we observe that discarding the nonparametric part of the model at prediction time hurts predictive accuracy significantly, thus confirming that the nonparametric fits of our method are useful where they are available.

\paragraph*{Ablation~2: Relevance of Variational EM.}
Our second ablation study goes one step further and removes the nonparametric part of the model already at training time.
As a result, the model can no longer be trained with variational EM and has to be trained with standard maximum likelihood estimation (MLE) instead (see ``Ablation Studies'' above).
We performed this ablation study only on the MoFo MCM model as performing the same ablation for the GNN MCM model would result in a simplified variant of the method by \citet{medina2022graph}, which we already compare to as part of our baselines (see \Cref{tab:comparison-medina} and ``Comparison to Baselines'').

The results for in-domain and out-of-domain predictions are labeled ``MoFo MCM (MLE)'' in \Cref{fig:main-results}.
For in-domain predictions, we observe that models trained with MLE perform significantly worse than models trained with variational EM, but better than if we train with variational EM and then discard the nonparametric part (see Ablation~1 above).
For out-of-domain predictions, the picture is less clear.
Here, models trained with MLE perform slightly better than their variational EM counterparts, in particular in terms of MSE, which penalizes outliers more strongly.
A possible explanation is that variational EM allows the parametric prior models to effectively ignore any mixture components that can be better modeled in a nonparametric way.
This would make the priors in variational EM less regularized, so they are more susceptible to overfitting to the training data, which can result in worse generalization to unseen mixture components in out-of-domain predictions.
However, this slight improvement of MoFo MCM (MLE) over MoFo MCM on out-of-domain predictions comes at the cost of significantly reduced performance on in-domain predictions, where the lack of a nonparametric model part prevents MoFo MCM (MLE) from specializing to components showing an anomalous behavior.
Further, the proposed GNN MCM model further improves performance over both MoFo MCM and MoFo MCM (MLE) significantly on both in-domain and out-of-domain predictions.

\begin{figure}[t]
  \centering
  \includegraphics[width=\columnwidth]{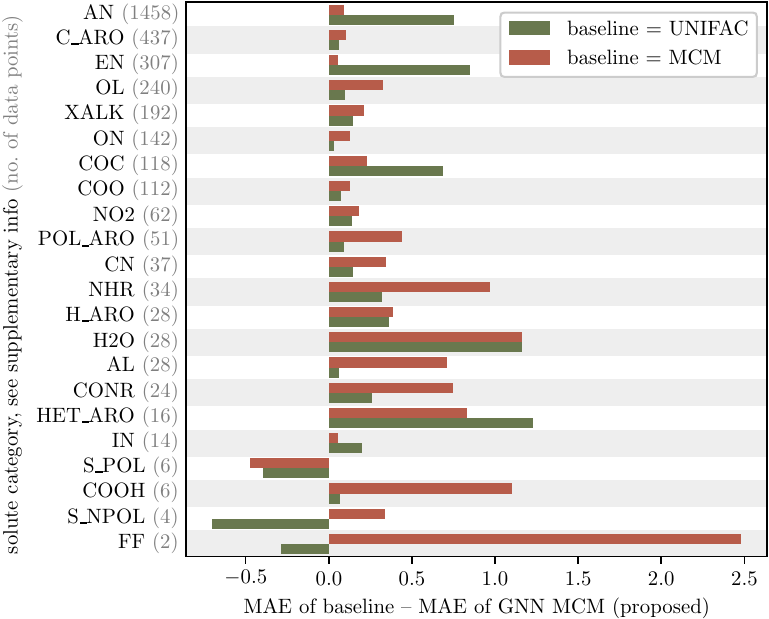}
  \caption{
  Improvement (in terms of mean average error, MAE) of the proposed GNN MCM method over UNIFAC and MCM, grouped by chemical category of the solute (see \Cref{tab:compound-categories} in the supplementary information for a definition of the categories).
  Our method consistently improves over both UNIFAC and MCM across almost all solute categories;
  we see regressions (negative improvements) only on three categories with poor statistics in this evaluation due to small sample sizes (see gray numbers).
  }
  \label{fig:solutes-by-group}
\end{figure}

\begin{figure}[t]
  \centering
  \includegraphics[width=\columnwidth]{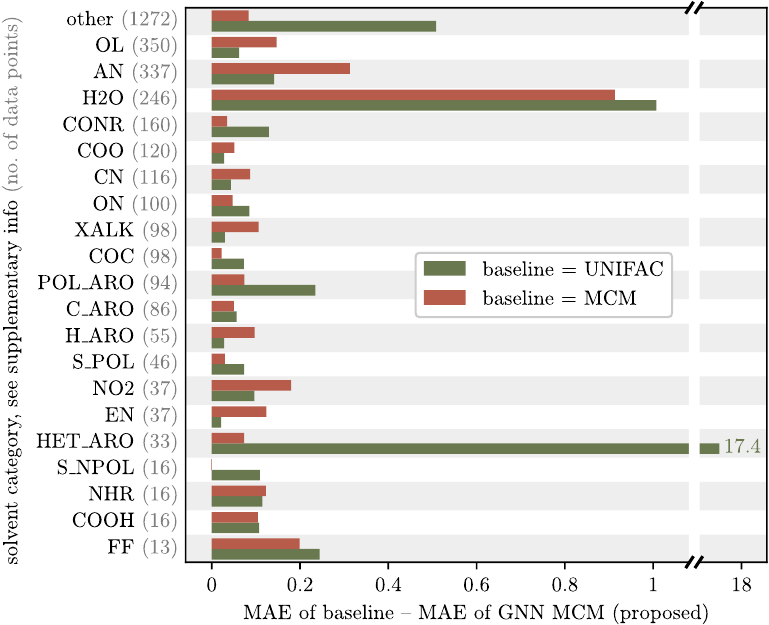}
  \caption{
  Improvement of the proposed GNN MCM over UNIFAC and over MCM, grouped by chemical category of the solvent (see \Cref{tab:compound-categories} in the supplementary information for a definition of the categories).
  Our method consistently improves over both UNIFAC and MCM across all solvent categories.
  }
  \label{fig:solvents-by-group}
\end{figure}

\paragraph*{Comparison by Chemical Structure.}
We analyze whether the improved predictive accuracy of our proposed GNN MCM method is systematic across all mixture types or limited to specific types of mixtures.
For this purpose, we manually assign each solute and solvent to a category based on its chemical structure, e.g., category ``XALK'' for substituted alkanes and alkenes, or category ``HET\_ARO'' for heteroaromatic compounds.
\Cref{fig:solutes-by-group} and \Cref{fig:solvents-by-group} show the improvement of our proposed GNN MCM over both UNIFAC (olive) and MCM (terracotta), grouped by the category of the solute and solvent, respectively.
We show in-domain predictions here so that we can compare to MCM.
Positive values in the figures indicate that GNN MCM has a lower mean average error (MAE) within the corresponding category than the baseline, whereas negative values indicate that the baseline performs better within a given category.

We find that the improvements of our proposed GNN MCM are systematic across almost all categories of solutes and solvents.
The only regressions occur within the solute categories ``S\_POL'' (strongly polar sulfurous compounds), ``S\_NPOL'' (weakly polar sulfurous compounds), and ``FF'' (perfluorinated compounds).
The results in these categories should be taken with a grain of salt as the dataset contains only very few mixtures that involve a solute from one of these categories (gray numbers in \Cref{fig:solutes-by-group}).
Thus, within these categories, the MAE averages only over 6, 4, or 2 values, respectively, making it highly susceptible to outliers.

\begin{figure}[t]
  \centering
  \includegraphics[width=\columnwidth]{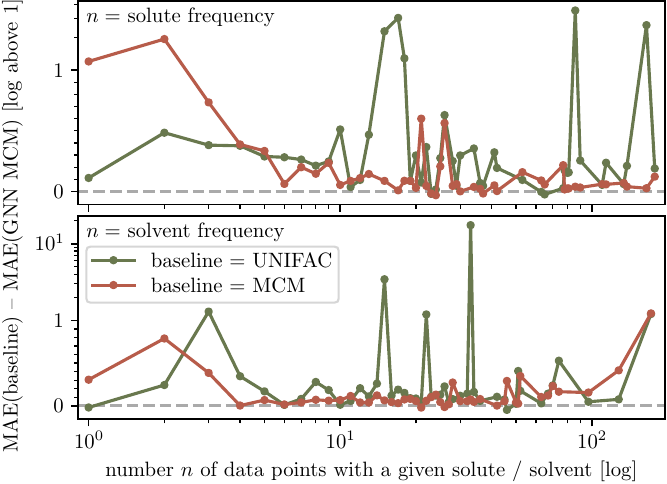}
  \caption{
  Improvement (in terms of mean average error, MAE) of the proposed GNN MCM method over UNIFAC and over MCM, grouped by the frequency~$n$ of solutes (top) or solvents (bottom) in the dataset.
  Our method consistently improves over both UNIFAC and MCM across both common and uncommon mixture components.
  Improvement over MCM (terracotta) is most pronounced for uncommon solutes and solvents (with small~$n$), which is as expected since MCM is fully nonparametric.
  }
  \label{fig:statistical-analysis}
\end{figure}

\paragraph*{Comparison by Data Availability.}
We finally analyze if the improvement in predictive accuracy for a given mixture depends on the amount of training data that is available for the two mixture components.
This analysis is motivated by the fact that our proposed GNN MCM trades off between a parametric and a nonparametric part, where parametric models tend to perform better in the low-data regime (because they can share statistical strength across structurally similar components) while nonparametric models tend to perform better in the high-data regime (because they can fit the data better due to fewer constrains).
We therefore investigate whether our method manages to use the best of both approaches in both regimes.

For each solute~$i$, we set $n(i)$ to the number of times that $i$ appears as a solute in the data set~$\mathcal D$.
The inverse of this function, $i(n):= \{i\in\mathcal D: n(i)=n\}$ maps each possible solute count~$n\in \mathbb N$ to all solutes in the data set that appear $n$ times in the dataset~$\mathcal D$.
For each~$n$ where $i(n)$ is not empty, we now calculate the average prediction error for all binary mixtures in the data set whose solute is in $i(n)$.
We proceed analogously for the solvents.

\Cref{fig:statistical-analysis} shows the improvement of our proposed GNN MCM over both UNIFAC (olive) and over MCM (terracotta) as a function of the frequency~$n$ of solutes (top) and solvents (bottom).
We observe that GNN MCM improves over both UNIFAC and MCM consistently across all solute and solvent frequencies~$n$ (i.e., almost all points in the plots lie above the dashed zero line).
The improvement over MCM (terracotta lines) is most pronounced in the regime of small~$n$, i.e., where few data points with the same solute or solvent exist.
This is to be expected since this is the regime where a fully nonparametric model like MCM tends to perform poorly.
The strong improvement for the solvent with highest~$n$ (right end of lower plot) can be explained as this solvent is water, for which a lot of data points with infrequent solutes exist in the dataset.

In summary, we find that our proposed method significantly improves predictive accuracy over both fully parametric and fully nonparametric baselines (\Cref{fig:main-results} and \Cref{tab:comparison-medina}), and that this improvement is consistent across mixture components from different chemical categories (\Cref{fig:solutes-by-group} and \Cref{fig:solvents-by-group}) and across varying amounts of training data for involved components (\Cref{fig:statistical-analysis}).

\section{Summary and Outlook}
In this work, we propose a method for predicting physico-chemical properties that combines a structure-based approach using graph neural networks (which are able to extrapolate across substances with similar chemical structure) with a representation-learning based approach (which allows the model to override structure-based predictions in anomalous cases).
The method significantly improves predictive accuracy over the state of the art in the studied problem of predicting activity coefficients in binary mixtures.
 
Our ablation studies identify the variational EM algorithm to be crucial for the success of the prediction method.
We think that variational EM can be a useful tool for many physico-chemical prediction problems since it balances structure-based and representation-learning based predictions by weighing off their respective uncertainties.

Future work should explore the application of our method to other properties such as diffusion coefficients or even fundamental quantities like interaction energies, which are at the core of established physical models of mixtures and based on which diverse mixture properties can be described.
In a broader context, our work provides additional evidence for the efficacy of graph neural networks for processing chemical structure information.
It would be interesting to study whether activations of hidden layers of the graph neural networks can be made interpretable to human domain experts, whether correlations between the hidden activations of different atoms can be used to identify relevant substructures of molecules, and whether such substructures correspond to the structural groups that are considered in established group contribution methods like UNIFAC.

\section*{Data and Software Availability}
We evaluate our methods on two datasets: a dataset that is licensed from the Dortmund Data Bank (DDB) \cite{onken1989thedortmund} and a dataset collected by \citet{brouwer2021trends}. 
The software packages for preprocessing and training our models are freely available. 
We provide the source code to replicate our results at \url{https://github.com/jzenn/gnn-mcm}.
For the comparison of our method to the model proposed by \citet{medina2022graph}, we directly use the files available from their GitHub repository.

\section*{Acknowledgements}
Johannes Zenn thanks Tim Xiao for helpful comments and discussions.
Funded by the Deutsche Forschungsgemeinschaft (DFG, German Research Foundation) under Germany’s Excellence Strategy – EXC number 2064/1 – Project number 390727645. 
This work was supported by the German Federal Ministry of Education and Research (BMBF): Tübingen AI Center, FKZ: 01IS18039A.
Robert Bamler acknowledges funding by the German Research Foundation (DFG) for project 448588364 of the Emmy Noether Program.
The authors would like to acknowledge support of the `Training Center Machine Learning, Tübingen' with grant number 01|S17054.
Fabian Jirasek gratefully acknowledges funding by DFG for project 528649696 of the Emmy Noether Program and for project 497201843 in the Priority Program Molecular Machine Learning (SPP 2363).
The authors thank the International Max Planck Research School for Intelligent Systems (IMPRS-IS) for supporting Johannes Zenn.



\balance


\bibliography{references}
\bibliographystyle{icml2024}

\newpage
\appendix
\onecolumn
\section{Supplementary Information}

\subsection{Model and Training Hyperparameters}

To find well-performing hyperparameters we employ sparse random grid searches (RGS) for all models we consider in this work.
We train $30$ models with hyperparameters sampled uniformly at random from a defined search space.
Each of the models is trained on the same training set and evaluated on the same test set. 
We pick the best model according to its mean squared error (MSE) on the test set.
To report fair error estimates, we exclude all data points of this test set from the 10-fold cross validation. 
Since the sample size of $30$ models is very small compared to the size of the search space, one might be able to further improve on our results by extending the RGS.

We schedule the learning rate either using a constant schedule or one out of nine schedules.
All schedules $S_i \in \mathcal S$ (except for the constant one) start from an initial learning rate $\epsilon$ that is decreased by a factor of (maximally) $10^{-1}$ (multiple times) during training.
After some initial warm-up steps $T_w$ the Robbins-Monro Learning Rate Scheduler \cite{robbins1951stochastic} computes the learning rate at epoch $t$ from an initial learning rate of $\epsilon_0$ as follows.
\begin{align}
    \epsilon_t
    &=
    \epsilon_0
    \;/\;
        \left(\frac{t - T_w}{b} + a\right)^\gamma
\end{align}
The three parameter combinations we consider are shown in \Cref{table:si:rm_scheduler}.
The Cyclical Learning Rate Scheduler \citep{smith2017cyclical} linearly increases the learning rate from an initial $\epsilon_- = 10^{-1}\epsilon_0$ to $\epsilon_+ = \epsilon_0$ and back to $\epsilon_-$. 
This is done $T_c$ number of times during training. 
We consider $T_c \in \{1, 2, 4\}$
The Step Learning Rate Scheduler decays an initial learning rate $\epsilon_0$ by $\gamma$ every $T_e$ epochs. 
Consequently, at epoch $t$ the learning rate equals $\epsilon_0^{\gamma\lfloor t / T_e \rfloor}$. 
We search over three combinations of parameters that are depicted in \Cref{table:si:step_scheduler}.
\begin{table}[!ht]
    \begin{center}
    \begin{tabular}{cccc}
    $T_w$ & $\gamma$ & $a$ & b \\
    \hline
    $1.5 \cdot 10^3$ & $0.5$ & $1.0$ & $150$ \\
    $1.5 \cdot 10^3$ & $0.6$ & $1.0$ & $300$ \\
    $1.5 \cdot 10^3$ & $0.8$ & $1.0$ & $900$
    \end{tabular}
    \end{center}
    \caption{Combinations of parameters for the RM scheduler.}
    \label{table:si:rm_scheduler}
\end{table}
\begin{table}[!ht]
    \begin{center}
    \begin{tabular}{cc}
    $\gamma$ & $T_e$ \\
    \hline
    $0.8$  & $1500$ \\
    $0.45$ & $3750$ \\
    $0.1$ & $7500$
    \end{tabular}
    \end{center}
    \caption{Combinations of parameters for the Step scheduler.}
    \label{table:si:step_scheduler}
\end{table}

The ELBO (that is maximized) can be formulated in various ways \cite{hoffman2016elbo}, two of which we include in the grid search: 
ELBO-KL maximizes the data log-likelihood under the variational distribution and simultaneously minimizes a KL divergence between the variational posterior distribution and the prior distribution. 
ELBO-Entropy maximizes the data log-likelihood as well as the log-prior under the variational distribution while simultaneously maximizing the entropy of the variational distribution. 

We apply skip connections (skip con. $\neq$ none) either every layer or every second layer.
Additionally in the GNN MCM model, we experiment with mean aggregation besides the sum aggregation.

\Cref{tab:hyperparameter_searches:fixed_hyperparameters_film_pmf} lists hyperparameter values we search over for the GNN MCM model.
\Cref{tab:hyperparameter_searches:fixed_hyperparameters_mofo_pmf} lists hyperparameter values we search over for the MoFo MCM model.
\Cref{tab:hyperparameter_searches:fixed_hyperparameters_mofo_mf} lists hyperparameter values we search over for the MoFo MCM (MLE) model.

\begin{table}[!ht]
\begin{center}
\begin{tabular}{ll}
    parameter & 
    value
    \\
    \hline
    loss & 
    ELBO-KL, ELBO-Entropy
    \\
    learning rate
    &
    0.005, 0.001, 0.0005, 0.0001
    \\
    $K$
    &
    4, 8, 16
    \\
    lr-scheduler
    &
    0, 1, 2, 3, 4, 5, 6, 7, 8 
    \\
    dropout prob.
    &
    $0.0$, $0.1$
    \\
    representation dim.
    &
    16, 32, 64, 128
    \\ 
    aggregation
    &
    sum, mean
    \\ 
    $L$
    &
    1, 2, 4, 6, 8
    \\ 
    skip con.
    &
    none, every \{1, 2\}-th layer
    \\
    bias
    &
    true, false
\end{tabular}
\end{center}
\caption{Hyperparameters of grid search for GNN MCM models.}
\label{tab:hyperparameter_searches:fixed_hyperparameters_film_pmf}
\end{table}

\begin{table}[!ht]
\begin{center}
\begin{tabular}{ll}
    parameter & 
    value
    \\
    \hline
    loss & 
    ELBO-KL, ELBO-Entropy
    \\
    learning rate
    &
    0.005, 0.001, 0.0005, 0.0001
    \\
    $K$
    &
    4, 8, 16
    \\
    lr-scheduler
    &
    0, 1, 2, 3, 4, 5, 6, 7, 8, 9
    \\
    dropout prob.
    &
    $0.0$, $0.1$
    \\
    representation dim.
    &
    16, 32, 64, 128
    \\ 
    $L$
    &
    1, 2, 4, 6, 8
    \\ 
    skip con.
    &
    none, every \{1, 2\}-th layer
\end{tabular}
\end{center}
\caption{Hyperparameters of grid search for MoFo MCM models.}
\label{tab:hyperparameter_searches:fixed_hyperparameters_mofo_pmf}
\end{table}

\begin{table}[!ht]
\begin{center}
\begin{tabular}{ll}
    parameter & 
    value
    \\
    \hline
    learning rate
    &
    $10^{-\{2,3,4\}}$,$5 \cdot 10^{-\{4,5\}}$
    \\
    $K$
    &
    $4, 8, 16$
    \\
    lr-scheduler
    &
    $0, 1, 2, 3, 4, 5, 6, 7, 8$
    \\
    dropout prob.
    &
    $0.0, 0.1$
    \\
    representation dim.
    &
    16, 32, 64, 128
    \\ 
    $L$
    &
    $1, 2, 4, 6, 8$
    \\ 
    skip con.
    &
    none, every $\{1, 2\}$-th layer
\end{tabular}
\end{center}
\caption{Hyperparameters of grid search for MoFo MCM (MLE) models.}
\label{tab:hyperparameter_searches:fixed_hyperparameters_mofo_mf}
\end{table}

\paragraph*{GNN MCM (in-domain)}
ELBO-Entropy, 0.005, 16, 0, 0.1, 64, sum, 2, none, true

\paragraph*{GNN MCM (out-of-domain)}
ELBO-Entropy, 0.0005, 8, 7, 0.0, 64, mean, 8, 1, true

\paragraph*{\citet{medina2022graph}}
ELBO-Entropy, 0.001, 16, 7, 0.1, 16, sum, 6, none, false

\paragraph*{MoFo MCM (in-domain)}
ELBO-Entropy, 0.0005, 8, 4, 0.1, 16, 1, none

\paragraph*{MoFo MCM (out-of-domain)}
ELBO-Entropy, 0.001, 8, 0, 0.1, 16, 6, 1

\paragraph*{MoFo MCM (MLE) (in-domain)}
$10^{-2}$, 8, 1, 0.0, 64, 8, none

\subsection{Solute and Solvent Categories}

\Cref{tab:compound-categories} defines the chemical categories used in the ``Comparison by Chemical Structure'' of the main text.
We assigned each solute and solvent to one of these categories manually based on their chemical structure formula using human expert knowledge.
These assignments were only used in the evaluation;
the model is unaware of our assignments.
Due to data licensing, we do not provide the solutes and solvents that fall into the chemical categories listed in \Cref{tab:compound-categories} but only provide the total number of compounds in each category.

\begin{table}[h]
\begin{center}
\begin{tabular}{@{\;}ll@{\;}}
    abbreviation & description \\
    \hline
    AN & alkanes (including compounds with long alkyl \\
        & groups $\geq 10$ C-atoms) \\
    EN & alkenes and dienes \\
    IN & alkynes \\
    C\_ARO & aromatic compounds without heteroatoms \\
    POL\_ARO & aromatic compounds with substituent heteroatoms, \\
        & pi-systems with inductive and mesomeric effects \\  
    H\_ARO & aromatic compounds with substituent heteroatoms \\
        & that can build stable hydrogen bonds \\
    HET\_ARO & heteroarenes \\
    XALK & chlorine, bromine, and iodine alkanes \\
    OL & alcohols \\
    COO & esters \\
    NHR & amines \\
    ON & ketones \\
    AL & aldehydes \\
    COC & ethers \\
    CN & nitrils \\
    NO2 & nitro compounds \\
    COOH & short carboxylic acids \\
    FF & perfluorinated compounds \\
    CONR & amides \\
    S\_NPOL & weakly polar sulfurous compounds \\
    S\_POL & strongly polar sulfurous compounds \\
    H2O & water and heavy water 
\end{tabular}
\end{center}
\caption{Manually assigned chemical categories for solutes and solvents.}
\label{tab:compound-categories}
\end{table}


\end{document}